\documentclass[a4paper,11pt]{article}
\usepackage{colacl,times,myexamples,tipa,amsmath,alltt,multirow,ulem}
\usepackage[dvips]{epsfig,rotating}

\renewcommand{\l}[1]{\normalem\emph{\textipa{#1}}}
\newcommand{\p}[1]{\textipa{#1}}
\def\sref#1{\S\ref{#1}}
\def\code#1{{\ttfamily\bfseries\scriptsize #1}}
\newcommand{\strich}{\rule{.99\linewidth}{.1pt}\\}
\newcommand{\startpiece}{\par\noindent\strich\vspace{-2.2\baselineskip}\footnotesize}
\newcommand{\stoppiece}{\vspace{-\baselineskip}\noindent\strich\par\noindent\normalsize}

\newcommand{\finallongpage}{\enlargethispage{\baselineskip}}

\title{Temiar Reduplication in One-Level Prosodic Morphology}
\author{Markus Walther \\ University of Marburg \\             FB09/IGS, Wilhelm-R{\"o}pke-Str. 6A, D-35032 Marburg, Germany \\
           \tt Markus.Walther@mailer.uni-marburg.de}
\begin{document}
\thispagestyle{plain}
\maketitle
\begin{abstract}                       
Temiar reduplication is a difficult piece of
prosodic morphology. This paper presents the first computational
analysis of Temiar reduplication, using the novel finite-state
approach of One-Level Prosodic Morphology
originally developed by Walther (1999b, 2000). After reviewing both the data and
the basic tenets of One-level Prosodic Morphology, the analysis
is laid out in some detail, using the notation of the FSA Utilities
finite-state toolkit (van Noord 1997). One important discovery is that
in this approach one can easily define a regular expression operator
which ambiguously scans a string in the left- or rightward direction
for a certain prosodic property. This yields an elegant account of
base-length-dependent triggering of reduplication as found in Temiar.  
\end{abstract}
\section{Introduction}
Temiar is an Austroasiatic language of the Mon-Khmer group spoken by
a variety of tribal people in West Malaysia \cite{benjamin:76}.  Its
intricate morphological system has received some attention in the
theoretical literature. The main focus has been on the aspectual
morphology of verbs, where an interesting pattern of partial reduplication
emerges that is sensitive to the size of the verbal root. For example,
in the active continuative, \l{gElg@l} `to eat' reduplicates both the
initial /g/ and the final /l/ of its monosyllabic base \l{g@l}.
In contrast, bisyllabic \l{s@luh} `to shoot' comes out as
\l{sEhluh}, where only the final /h/ is copied, this time as an infix.

Temiar reduplication thus appears to be a suitably rich testing ground for a novel
approach to reduplication developed by \cite{walther:99,walther:00}
within a finite-state framework. Even though that 
approach, One-Level Prosodic Morphology, was presented from the outset
as being generally applicable, it has been 
proven time and time again that only concrete empirical application of
a particular approach to computational morphology and phonology will fully reveal its
inherent virtues and weaknesses. As an example, \cite{beesley:98}
reports that it was actual experimentation with grammars of word-formation in
Arabic and Hungarian which fully revealed the negative effects of modelling
long-distance circumfixional dependencies in purely finite-state
terms, subsequently leading to some suggestions for improvement.

It is perhaps worth emphasizing that \cite{walther:99}'s solution for
reduplication in a finite-state context  is preferrable for
cross-linguistic validation precisely because it is the first that solves
the problem in the {\em general} case. Because reduplication often 
involves copying of a strictly bounded amount of material, the bounded
case {\em could} in principle be modelled as a finite-state process 
by enumerating all possible forms of the copy and then making sure
each was matched to the proper stem. To solve this simplified problem, no new
techniques are needed in theory. In practice however, the brute-force
enumeration approach apparently has not been pursued further, apart from
isolated examples (see Antworth \shortcite{antworth:90}, p.157f for a
fixed-size case in Tagalog). This is probably because such an approach
is awkward to specify in actual grammars and because it will
inevitably lead to an explosion of the state space (Sproat
\shortcite{sproat:92}, p.161). Finally and in contrast to
\cite{walther:99},  it would clearly break down for {\em   productive} total
reduplication, which is isomorphic to the context-sensitive language
$\{ww|w \in \Sigma^{+}\}$. 

A second motivation for choosing Temiar is that all prior analyses of its data are
heavily underformalized and incomplete, irrespective of whether they
are situated in the older rule paradigm
\cite{mccarthy:82a,broselow.mccarthy:83,sloan:88,shaw:93} or an  
optimality-theoretic setting
\cite{gafos:95,gafos:96,gafos:98a,gafos:98b}. Hence a formalized and
computationally tested analysis that strives to keep a healthy balance
with respect to linguistic adequacy would represent significant progress on its own. 

In the rest of the paper I will attempt to provide just such an
analysis, beginning in \sref{tem} with a presentation of the relevant
data. Next, section \sref{olpm}  reviews the core of One-Level
Prosodic Morphology, which will be used as formal background. Using
that background, the analysis is then fully developed in
\sref{ana}. The paper concludes with some discussion in \sref{conc}.
\section{\label{tem}Temiar reduplication}
All data on Temiar reduplication in this section come from
\cite{benjamin:76}, the main source on the subject.%
\footnote{We will abbreviate further references to this work with ``(B: $<$page
  number$>$)'' in the text. Moreover, to highlight reduplicated parts in the data
  they will often be printed in bold.}
According to Benjamin, the characteristic aspectual paradigms of
``monosyllabic and schewa-form verbs'' (B:168) are as follows (B:169):
\begin{examples} \item \label{para}
\begin{tabular}[t]{@{}ll|ll@{}}
 & \fbox{\footnotesize `to call'} & \multicolumn{2}{l}{\fbox{\footnotesize `to lie down/sleep/marry'}}
 \\[1ex]
 & `monosyllabic' & \multicolumn{2}{l}{`schewa-form'}\\ \hline
\multirow{3}{1mm}{\tiny \shortstack{a \\ c \\ t \\ i \\ v \\e}}
 & \p{"kOOw} & \p{s@."lOg} & \footnotesize perfective \\
 & \p{{\bf k}a."kOOw} & \p{sa."lOg} & \footnotesize simulfactive \\
 & \p{{\bf k}E{\bf w}."kOOw} & \p{sE{\bf g}."lOg} & \footnotesize continuative \\ \hline
\multirow{3}{1mm}{\tiny \shortstack{ {} \\ c \\ a \\ u \\ s \\ a \\ t\smash{.}}}
 & \p{tEr."kOOw} & \p{sEr."lOg} & \footnotesize perfective \\
 & \p{t@.ra."kOOw} & \p{s@.ra."lOg} & \footnotesize simulfactive \\
 & \p{t@.rE{\bf w}."kOOw} & \p{s@.rE{\bf g}."lOg} & \footnotesize continuative 
\end{tabular}
\end{examples}
We have inferred syllabifications in \ref{para} from the statement that
``only two types of syllables occur: {\em open syllables} of canonical
form CV, and {\em closed syllables} of canonical form CVC''
(B:141). Note that Benjamin abstracts from vowel length here. %
Word-level stress, which is ``falling
regularly on the final syllable'' (B:139), is likewise inferred in
\ref{para}. Observe that only monosyllabic 
roots like \l{kOOw} reduplicate their initial consonant in the
non-perfective aspectual forms of the active, while longer roots like
\l{s@lOg} do not. This contrasts with obligatory reduplication of the
root-final consonant in the continuative. 

An important further generalization is that all extra segmental
material beyond the bare root is inserted immediately before the
stressed syllable, leading to prefixation for monosyllabic roots, but
infixation in polysyllabic ones \cite{gafos:98a}. From this point of
view we can also see a correlation between 
the fact that causative forms of monosyllabic roots -- which must be at
least bisyllabic -- begin with a fixed
/t/%
\footnote{Or /b/, if the root starts in /c,t/: /\p{caaP}/ `to eat'
  gives /\p{bEr.caaP}/ `to feed' (B:169).}
 and the restriction that words must ``always begin and end with a
consonant'' (B:141). In triconsonantal roots like \l{s@lOg} that restriction is taken
care of by the first root consonant itself, so no fixed segment needs
to appear.

According to Benjamin, prefinal syllables -- which are unstressed --
can show alternation of their vocalic quality: ``In prefinal closed
syllables the inner vowels /e \p{@} o/ are replaced by the outer
vowels /i \p{E} u/ respectively'' (B:144). This descriptive generalization accounts for the
remaining contrasts in \ref{para}, witness e.g.\/ \l{s@.lOg} versus \l{sEg.lOg}.

It is interesting to see that Temiar even exhibits phonological
modifications between base and reduplicant, affecting consonants in
the continuative:
\begin{examples} \item \label{mod}
\begin{tabular}[t]{@{}l@{}cl@{}r@{}}
\p{yaap} & $\rightarrow$ & \p{{\bf y}E{\bf m}.yaap} & {\footnotesize `to cry' (B:143)} \\
\p{p@t} & $\rightarrow$ & \p{{\bf p}E{\bf n}.p@t} & {\footnotesize `to long for' (B:146)} \\
\p{s@.lOOk} & $\rightarrow$ & \p{sE}{\bf\p{N}}\p{.lOOk}  &{\footnotesize `to hunt
successfully' (B:146)} 
\end{tabular}
\end{examples}
Benjamin explains that medial coda consonants from the class of
oral voiceless stops turn into their voiced nasal equivalents in
Northern Temiar (and to plain voiced stops in the Southern dialect; B:143).

It is of some importance to clarify a number of further aspects of the
data and their interpretation. First, theorists have frequently  employed the stronger term `minor
syllables' for Benjamin's prefinal syllables, reflecting their alleged
special status by means of an impoverished 
representation (e.g.\/ empty syllable nuclei in \cite{gafos:98a}) and/or
further formal mechanisms (e.g.\/ a ban on full vowels in prefinal
position {\sc *Prefinal-V} \cite{gafos:98b}). We do not follow this
move here, because empirically it is neither true that penultimate
vowels are categorically restricted to schwa-like vowels (\l{halab}
`to go downriver', \l{sindul} `to float', etc.) nor are there any
solid statistics of a presumed tendency to vowel reduction in
unstressed syllables, nor can the variable quality of prefinal vowels
be consistently derived from flanking consonants. Hence, such
penultimative vowels are to be lexically specified as alternating.

Second, Benjamin's subclass restriction of \ref{para} to ``monosyllabic and
schewa-form verbs'' correctly excludes polysyllabic roots like the
already mentioned \l{halab} and \l{sindul}, where prefinal open
syllables with vowels outside of /\p{e @ o}/ occur. These roots
undergo ``very few morphological changes'' (B:170), basically 
proclitization.

Third, paradigms for a given root are hardly ever complete,
with various irregularities and non-productive patterns also
occuring (B:169f). Again,  a good deal of lexicalization would seem necessary
to correctly describe Temiar verbs in a realistic grammar fragment.

Given this descriptive summary, our goals for the upcoming analysis
are, first, to treat the {\em full} paradigm of \ref{para}. As a second
goal, we would like to reflect the emergent formal desiderata in a   
 transparent way, in particular referring to the need to account for {\em repetition,
    truncation, infixation} and {\em phonological modification}.
Thirdly, we will attempt a {\em compositional} analysis of the morphological
  exponency of aspect.

\section{\label{olpm}One-Level Prosodic Morphology}
In order to provide the necessary background for the Temiar analysis
in \sref{ana}, this section briefly reviews the finite-state approach 
to prosodic morphology developed in \cite{walther:99},

That work itself was presented as an extension to
\cite{bird.ellison:94}'s One-Level Phonology framework, where
phonological representations, morphemes and more abstract
generalizations are all finite-state automata that express
surface-true constraints on word forms, and constraint combination is
by automata intersection.

In a nutshell, the extension comprises three main components. We
(i)  represent phonological strings differently for purposes of
modelling prosodic morphology, (ii) implement
reduplicative coyping by automata intersection, and (iii) introduce
a resource-conscious variant of automata.

\begin{figure}[htb]
\epsfig{file=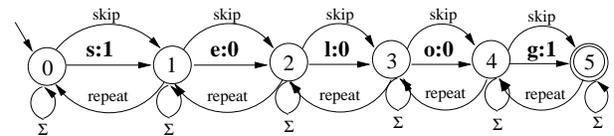,width=\linewidth}
\caption{\small add\_repeats(add\_skips(add\_self\_loops(selog)))}
\label{selog}
\end{figure}

For (i), operators are provided that construct enriched automata
from a simple string automaton, in particular giving it a kind of
doubly-linked structure so that the symbol repetition inherent in reduplication
translates into following backwards-pointing technical transitions.
The individual enrichments involve only local computation per state or 
transition, so that on-the-fly implementation is easy if desired.
In other words, one does not necessarily have to enrich the entire
lexicon in advance.
\paragraph{Enriched representations} In a bit more detail, the
enrichments of (i) are as follows. The three  
aspects of {\em reduplication} or symbol repetition, {\em truncation}
or symbol skipping and {\em infixation} or transitive, non-%
immediate precedence of symbols are reflected in three regular
expression operators, $add\_repeats, add\_skips,
add\_self\_loops$. Each takes the underlying automaton $A$ of a
regular language $L_A$ as its only argument. Formally, they can be defined
as follows:
\begin{examples}
\item 
 Let $A = (Q, \Sigma, \delta, q_0, F)$ be the minimal $\epsilon$-free%
\footnote{Minimality prevents non-(co)-accessible transitions from
  getting enriched, while lack of $\epsilon$ transitions keeps
  positional $skip/repeat$ `movement' in lockstep with segmental positions.}
finite-state automaton for $L_A$, with $Q$ a finite set of
states, finite alphabet $\Sigma$,  transition function $\delta :
Q\times\Sigma\mapsto 2^Q$, start state $q_0\in Q$ and set
of final states $F\subseteq Q$. 

\begin{examples}
\item Assume $repeat \not\in\Sigma$. \\ $add\_repeats(A) \stackrel{def}{=} (Q, \Sigma',
\delta', q_0, F)$, where  $\Sigma' = \Sigma \cup \{repeat\}$,\\ $\forall x
\in \Sigma\,\forall q\in Q{:}\,\, \delta'(q,x) = \delta(q,x)$ and\\
$\forall p\in Q{:}\delta'(p,repeat)=\{q|\, p\in\delta(q,x)\}$ 

\item Assume $skip \not\in\Sigma$. \\ $add\_skips(A) \stackrel{def}{=} (Q, \Sigma',
\delta', q_0, F)$, where  $\Sigma' = \Sigma \cup \{skip\}$, \\ $\forall x
\in \Sigma\,\forall q\in Q{:}\,\, \delta'(q,x) = \delta(q,x)$ and\\ 
$\forall q\in Q{:}\,\, \delta'(q,skip) = \delta(q,x)$

\item \label{ads}$add\_self\_loops(A) \stackrel{def}{=} (Q, \Sigma,
\delta', q_0, F)$, where \\ $\delta' = \delta \cup
\{(q,\sigma,\{q\}) |\, q\in Q,\, \sigma \in \Sigma\}$
\end{examples}
\end{examples}
An example enrichment of Temiar \l{s@lOg} is shown in figure
\ref{selog}. One can imagine how $skip$ and $repeat$ transitions
allow, figuratively speaking, forward and backward movement within a
string, while self loops will absorb infixal morphemes that are
intersected with fig.\/ \ref{selog}. Finally, so-called {\em
  synchronization bits} {\bf :1, :0} were introduced in
\cite{walther:99} to define the extent of a reduplicative base
constituent in a segment-independent way.
Bit value {\bf :1} marks the edges and {\bf :0} the interior segments of a base, as shown in fig.\/
\ref{selog} for a hypothetical whole-root reduplication pattern.
In actual practive, synchronization bits are sets of symbols, just like the rest 
of the alphabet. Sets as transition labels improve over traditional automata in
terms of automata compactness, were already proposed for phonology in
\cite{bird.ellison:92} and do not increase mathematical expressivity
beyond regular languages.%
\footnote{Of course, the identity requirement for matching transitions
in traditional automata intersection must be replaced by a non-empty intersection
requirement for set-based matching.} 
Hence, the segmental part of fig.\/ \ref{selog} may
be defined in a modular fashion through the intersection of strings of symbol sets that mention
only certain dimensions (here: phonemes and synchronisation bits),
being underspecified for the unmentioned dimensions. We will again follow
\cite{walther:99} in conceiving of sets as types arranged in a type
hierarchy that is structured by set inclusion, and also in allowing arbitrary boolean
combinations of types.

\paragraph{Copying as intersection} Given enriched representations as
in fig.\/ \ref{selog}, various 
patterns of reduplication are now easy to define. We can denote a
synchronised abstract string by the regular expression \[ base\equiv seg{:}1\,
seg{:}0^*\,seg{:}1\] where $seg$ is the type subsuming all phonological
segments. Then hypothetical total reduplication -- unattested in
Temiar, but wellknown from Indonesian and many other languages --
is described by \[total \equiv base\, \boldsymbol{repeat}^*\, base\] A variant slightly
more akin to Temiar -- and actually attested in the neighbouring
language Semai -- that $skip$s the interior of the base in a prefixed
reduplicant is just as easy: \[semai \equiv  seg{:}1\,\boldsymbol{skip}^*\, seg{:}1\, \boldsymbol{repeat}^*\, base\]
 
Ignoring self loops for the moment, all we need now to apply a reduplication
pattern to an enriched base representation is simply to {\em intersect}
the former with the latter: automata intersection has sufficient
formal power to implement reduplicative copying! Here is an example,
using the abbreviation $selog\equiv \boldsymbol{s{:}1\,e{:}0\,l{:}0\,o{:}0\,g{:}1}$ 
for perspicuous display:
{\small \[add\_repeats(selog)\, \cap\, total \equiv
selog\,\boldsymbol{repeat^5}\,selog\]}
As pointed out in \cite{walther:00}, generalizing to a {\em set} of bases 
involves nothing more than enriching each base separately, then
forming the union of the resulting automata. The opposite order
would produce unwanted cross-string repetition, since $add\_repeats$ does
not distribute over union. However, an unpublished experiment shows that 
on-demand implementation of a slightly modified $add\_repeats$ can help to
preserve the memory efficiency of building a minimized base lexicon as the
union of individual base strings first. Due to lack of space, the
details will be reported elsewhere.

\paragraph{Resource consciousness}
As much as we need the formal means provided by self loops for
infixations like Temiar \l{s-a-lOg}, the resulting automata overgenerate
massively. What's missing according to \cite{walther:99} is a
distinction between explicitly contributed, independent information (e.g.\/ the
infix \l{-a-} itself) and contextual, dependent information that is tolerated but must
be provided by other constraints (e.g.\/ the $1 \stackrel{\Sigma}{\rightarrow}1$ self loop
that {\em hosts} the infix).
Therefore, a parallel distinction between two kinds of
symbols -- producers and consumers -- was introduced. In
that scenario a symbol represents  an information resource that needs
to be produced at least once, then can be consumed arbitrarily
often. To utilize the distinction, an additional P/C bit accompanies
symbols, with P/C = 1 for producers. All symbols introduced by the three enrichment operators are consumers. Furthermore,
automata intersection is made aware of these resource-conscious
notions by splitting it into two variants: In open interpretation
mode, P/C bits of matching symbols are combined by logical OR, so
that a result transition will be marked as a producer whenever at least 
one argument transition is a producer. In closed interpretation
mode, combination is by logical AND instead, allowing only
producer-producer matches. Grammatical evaluation can then be
characterized as follows: \[ \scriptsize (\text{Lexicon}\cap_{open}\text{Constraint}_1\dots\cap_{open}%
\text{Constraint}_N)  \cap_{closed}\boldsymbol{\Sigma}^{*} \]
Here and elsewhere, producers are in bold print. Note the final
intersection with the universal producer language, which eliminates
unused consumer transitions, the main source of overgeneration.

\section{\label{ana}The analysis}
We have assembled enough background now to proceed to the actual
analysis of the Temiar data in \ref{para}. The analysis is implemented 
using FSA Utilities, a finite-state toolbox written in Prolog which
encourages rapid prototyping \cite{vannoord:97}. Figure \ref{fsa}
shows a relevant fragment of its syntax (extensions and modifications
in italics).
\begin{figure}[htb]
\begin{tabular}[t]{cl}
\tt \verb+{}+ & empty language \\
\tt [E1,E2,\dots,En]    & concatenation \\
\tt \verb+{+E1,E2,\dots,En\verb+}+    & union\\
\tt E*                  & Kleene closure\\
\tt E\verb+^+                  & optionality\\
\tt E1 \verb+&+ E2      & intersection\\
\it\tt  A $\underset{{-}{r-}>}{^{{-}{l-}>}}$ ( B / C) & monotonic rules \\
$\sim$\it\tt S &  set complement \\
\it  Head(arg1, \dots, argN) := Body & macro def.
\end{tabular}
\caption{Regular expression operators}\label{fsa}
\end{figure}

In displaying the grammar, we will take liberty in suppressing
certain definitions in the interest of conciseness, relying on the
mnemonic value of their names instead. A case in point is
\code{producer(T), consumer(T)}: since the names are self-explanatory,
it suffices to note that the only argument \code{T} contains type
formulae that denote the symbol sets, as explained before. Allowable
type-combining operators are conjunction \code{\&}, disjunction \code{;} 
and negation $\boldsymbol{\sim}$. The same
goes for monotonic rules, which -- unlike rewrite rules -- can only specialize their focussed segmental 
position \code{A} to \code{B}. They exist in two variants, where \code{A -r-> B/C} notates
the case where context \code{C} is right-adjacent to the focus ($A
\rightarrow B / \underline{\phantom{X}}\, C$), and vice versa for \code{A -l-> B/C}.
\paragraph{Syllabification} 
To define the reduplicant in prosodic terms later on, we need
\code{syllabification} in the first place. Here a simplified finite-state
version of a proposal by \cite{walther:99a} is employed. Its key idea 
is to allow incremental assignment of syllable roles to segmental
positions via a featural decomposition of
the three traditional roles, using two binary-valued features \code{ons} and \code{cod}: 
\begin{examples}
\item {\footnotesize
\begin{tabular}[t]{@{}|l||r|r|} \hline 
{\bf O}nset & \code{ons} & $\sim$\code{cod} \\
{\bf N}ucleus & $\sim$\code{ons} & $\sim$\code{cod} \\
{\bf C}oda & $\sim$\code{ons} & \code{cod} \\
{\bf C}oda{\bf O}nset & \code{ons} & \code{cod} \\ \hline
\end{tabular}}
\end{examples}
As a side-effect, one gets the fourth role {\footnotesize\bf CO}, a monosegmental prosodic
representation of true geminates. The subcomponent \code{sbs}, for
\underline{s}onority-{\underline b}ased \underline{s}yllabification, itself
rests on the computation of \code{sonority\_differences} between adjacent
segmental positions (not shown), where sonority may either go
\code{up} or \code{down}. Together with some self-explanatory
constraints \code{obligatory\_wordinternal\_onsets} and \code{no\_geminates},
prosodic surface wellformedness is then welldefined.
Only \code{if\_doubly\_synced\_edge\_then\_stressed} may seem slightly 
odd, since it has a purely technical character: it rules out certain illformed alternatives in
wordforms. Note, however, that the necesssity of such technical constraints, which are
certainly implicit in informal analyses as well, can only be reliably 
detected in computerized analyses such as the present one, which
allow for mechanical enumeration of a grammar's denotation.

\startpiece\begin{verbatim}

sbs := [ {  [consumer(down&~ons), 
             consumer(segment&~'Nuc')],
            [consumer(up&~'Nuc'), 
             consumer(segment&~cod)
         } *, no_final_onset ^].

no_initial_coda := consumer(segment&~cod).
no_final_onset := consumer(segment&~ons).

syllabification := sonority_differences&
 sbs&[no_initial_coda, sbs].

% -- further constraints ---
obligatory_wordinternal_onsets := 
  ( segment -r-> ons / 'Nuc' ). % _ 'N'

no_geminates := consumer(~'CO')*.

prosodic_constraints := obligatory_word-
 internal_onsets & no_geminates &
 if_doubly_synced_edge_then_stressed.

if_doubly_synced_edge_then_stressed :=
 [( {consumer(~':1'),
    [consumer(':1'),consumer(~':1')],
    [consumer(':1'),consumer(':1'),
     consumer(stressed)]
    } *), consumer(':1') ^].
\end{verbatim}\stoppiece%
\vspace*{-2\baselineskip}\paragraph{Stress} Given the assignment of syllable roles to segmental 
positions, we are now ready to define Temiar word \code{stress}. 
A possibly empty sequence of \code{prefinal\_syllables}, each of which is
constrained to be of shape $O N (C)$ and \code{unstressed}, is followed by a final
\code{stressed syllable}. The macro \code{ends\_before\_last\_syll} makes sure
that the dividing line between the penultimate and ultimate syllable is
drawn correctly.\finallongpage

\startpiece\begin{verbatim}

stress := [prefinal_syllables &
            ends_before_last_syll, 
            syllable].

prefinal_syllables := 
   ([consumer('Ons'), consumer('Nuc'), 
    (consumer('Cod') ^) ]*) &
    consumer(unstressed)*.

ends_before_last_syll:=([consumer(segment)*,
                 consumer(segment&~ons)]^).

syllable := [consumer(ons)+,consumer('Nuc'), 
             consumer(cod)*] &  
            (consumer(stressed)*).
\end{verbatim}\stoppiece
\paragraph{Stems} We proceed towards the definition of a \code{stem}
by noting that --  as described in \sref{tem} -- both the extent of a
\code{base}'s phonological material {\em and} its stress pattern are necessary
prior knowledge for adding aspectual morphemes in the appropriate
way. Hence, we impose the respective constraints onto the {\em isolated base
string} in \code{stem0}, before wrapping the result in the usual
enrichments. However, the addition of self loops for
infixation this time is {\em a priori} restricted to the position immediately before
a stressed onset, in accordance with the descriptive generalization
stated in \sref{tem}. Experiments have shown that using the unrestricted $add\_self\_loops$ of
\ref{ads} would cause much unnecessary hassle in {\em a posteriori}
restriction of the possible infix locations to the actually attested
ones. It thus appears that Temiar provides a first case for 
further parametrization of at least one of the original operators from 
\cite{walther:99}:
\startpiece\begin{verbatim}

base := [consumer(':1'),consumer(':0')*,
         consumer(':1')].

stem0(StemMaterial) := 
 add_self_loop_before(stressed&'Ons', 
  add_repeats(add_skips(StemMaterial &
   base & syllabification &
   prosodic_constraints & stress))).

stem(Segments) := 
 stem0(stringToSegments(Segments)).
\end{verbatim}\stoppiece
Definitions for the actual stem entries of \code{selog, koow, yaap} are shown 
below, using the ASCII-IPA mapping {\tt \{@ $\mapsto$ \p{@}, E
  $\mapsto$ \p{E}, O $\mapsto$ \p{O}\}}. In evaluating the first
entry, the schwa actually 
translates into a producer-type disjunction ({\bf \p{@};\p{E}}) with the help of
\code{stringToSegments}. It thus makes sense to constrain this free
alternation further, which is the purpose of
\code{has\_prefinal\_syllable}. While the monosyllable \code{koow}
needs no extra treatment, \code{yaap} is an example of a stem ending
in an \code{alternating\_labial}, whose definition however is
straightforward (\code{medial, final} refer to a positional
classification of the word that is defined later):\enlargethispage{2\baselineskip}
\startpiece\begin{verbatim}

selog := stem("s@lOg") &
       has_prefinal_syllable.

koow := stem("kOOw").
yaap := stem0([stringToSegments("yaa"), 
               alternating_labial]).

alternating_labial := {producer(p&final), 
                  producer(m&medial&cod)}.
\end{verbatim}\stoppiece
If we now define \code{has\_prefinal\_syllable} itself, we have
completed the components that make up \code{stem}. 
While the definition really targets the prefinal vowel, its preceding
onset and the stretch of arbitrary material after it must also be
mentioned. To tolerate interspersed \code{technical\_symbols}, the
\code{ignore} operator is used \cite{kaplan.kay:94}.

The purpose of \code{prefinal\_V} is to
control the alternation between `outer' and `inner' vowel, here
parametrized for \l{E}$\sim$\l{@} only. It does so by referencing the next
syllable role: if it is consistent with \code{ons}, that vowel resides 
in an open syllable, hence the \code{close\_mid} variant (\l{@}) will be
selected. Two elsewhere cases deal with closed syllables and the
possible presence of a technical symbol:
\startpiece\begin{verbatim}

has_prefinal_syllable :=
  ignore([consumer('Ons'), 
          prefinal_V(('E';'@'), 
              ':0'&unstressed),
          consumer(anything) *], 
          technical_symbols).

technical_symbols := 
 (consumer((skip;repeat)) *).

prefinal_V(Quality, Common) :=
 { [producer(Quality&close_mid&Common), 
    consumer(ons)],
   [producer(Quality&~close_mid&Common), 
    consumer(cod)],
   [consumer((skip;repeat))]
 } ).
\end{verbatim}\stoppiece
\paragraph{Aspectual affixes} It is time to concentrate on the
 most interesting part, and that is how to define the affixes.
Again the general picture will be to see them as constraints on word
forms which are imposed by intersection. We begin with the
\code{simulfactive}. The claim here is that its characteristic pattern is the realization
of the initial  base segment (\code{:1}), followed by the infixed melodic element
/a/, and then the entire string that begins with the stressed
onset. Phrasing the pattern this way already suffices to capture
the difference in reduplication behaviour between \l{"kOOw} and \l{s@"lOg}:
if we have inserted the \l{-a-} after the initial consonant in the first base,
the stressed onset is {\em to the left of /a/'s position}, whereas in
the second base that onset is found {\em to the right}. Thus,
repetition of segments is necessary to avoid ungrammaticality due to
constraint violation in the first case (\l{k-a-"kOOw}), but not in the
second (\l{s-a-"lOg}).

This behaviour is most naturally modelled by defining a new operator
\code{seek(X)}, which allows for ambiguous movement {\em either} to
the left (\code{repeat}) {\em or} to the right (\code{skip}) before
imposing the restriction \code{X}. This operator is applied to infixal 
/a/ because it is precisely the infix which needs to `seek' its
prosodically defined unique insertion point, i.e.\/ self loop.
Finally, to ensure that the other aspectual morphemes can play their part later
on, the entire pattern is wrapped in \code{align} to tolerate further
material before (\code{align\_right}) and after it (\code{align\_left}):
\startpiece\begin{verbatim}

simulfactive := 
  align([consumer(':1'),
         seek([producer(a&':0'&unstressed),
         consumer(stressed&'Ons')])]).

seek(X) := 
 [{producer(skip)*,producer(repeat)*},X].

align_left(X):=[X,consumer(anything)*].
align_right(X):=[consumer(anything)*,X].
align(X) := align_right(align_left(X)).
\end{verbatim}\stoppiece
Moving on to the \code{continuative}, we can see that the relevant
formal generalization is a bit more complex. Again we start off with
the initial base segment (\code{:1}), but then seek a place to infix
the constant /\p{E}/, before we \code{skip\_to} the next synchronised
base position (\code{:1}), which inevitably will be the final one. The 
pattern is completed by again seeking the stressed onset, from which
realization of the string proceeds uninterrupted due to the licensing
of extra material that the \code{align} wrapper provides. This
produces a similar contrast with respect to (non-)reduplication of the first base
position, but makes both the repetition of the last base segment and
the \sout{truncation} of its interior material obligatory in both base types
\mbox{(\l{k-E-} \sout{\it oo} \l{w"kOOw} vs. \l{s-E-} \sout{\it lo} \l{g"lOg}):}
\startpiece\begin{verbatim}

continuative := 
 align([consumer(':1'),
    seek([producer('E'&':0'&unstressed)]),
       skip_to(consumer(':1')),
       seek(consumer(stressed&'Ons'))]).

skip_to(X) := [producer(skip)+, X]. 
\end{verbatim}\stoppiece
What is left now is the proper definition of the \code{causative}.
Here we observe from \ref{para} that the causative morphology always starts
word-initial, hence the use of  \code{align\_left}. We have a default
consonant /t/ whose realization we must somehow force in the
monosyllabic roots. Next comes a vowel, whose quality -- \p{@} or
\p{E} -- is again regulated by
the familiar \code{has\_prefinal\_syllable}. Finally, the characteristic
fixed element /r/ is specified. Upon second thought, the /t/ is
guaranteed to appear in monosyllable roots, because prefinal syllables
always require an onset. The default absence of the /t/ -- when
not needed on prosodic grounds -- is again encoded by the
producer/consumer distinction, which contrasts the two disjuncts of the
parametrized macro \code{default}:
\startpiece\begin{verbatim}

causative := 
 align_left([default(t&unstressed,':1'), 
             producer(vowel),
             producer(r&':1'&unstressed)])&
            has_prefinal_syllable.

default(Optional, Common) := 
  { producer(Common&Optional), 
    consumer(Common) }.
\end{verbatim}\stoppiece
\paragraph{Entire words} We can put the pieces together now  by
first defining the \code{word} constraint as the conjunction of 
syllabification and related prosodic constraints plus a classification 
of the word's segmental positions into \code{initial,medial,final} ones. Again,
this is modulo interspersed $repeat$ or $skip$ symbols. This actually
means that base syllabification and word syllabification must match
up, but fortunately this is indeed a property of our Temiar data.

\begin{figure*}
   \centering
   \fbox{\begin{minipage}{\linewidth}
\epsfig{file=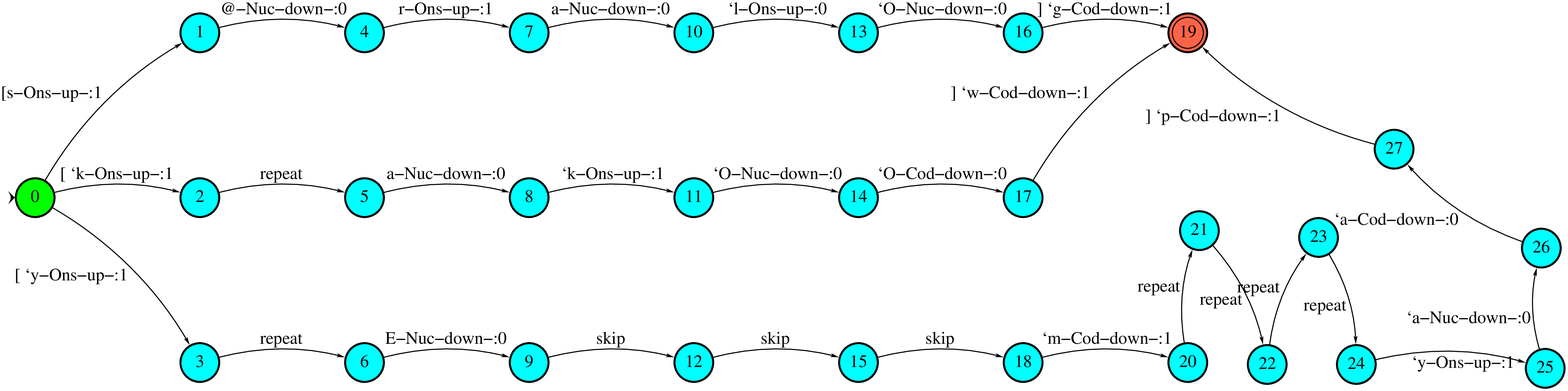,width=\linewidth}
\caption{Temiar reduplications \l{s@ralOg}, \l{kakOOw}, \l{yEmyaap}}\label{forms}
\end{minipage}}
\end{figure*}

Second, \code{wordform} conjoins the previous constraint
with its parameter \code{X} -- which will contain the conjunction of
stem and aspect morphemes --, before eliminating leftover consumer
symbols with the help of \code{closed\_interpretation}:
\startpiece\begin{verbatim}

word := ignore(syllabification &
               prosodic_constraints &
               positional_classification,
               technical_symbols).

positional_classification := 
[consumer(initial),consumer(medial)*,
 consumer(final)].

wordform(X):=closed_interpretation(X&word).
\end{verbatim}\stoppiece
These definitions have removed the last barrier to evaluating
expressions like \code{wordform(selog \& simulfactive \& causative)} or
even suitable disjunctive combinations of such expressions which
define entire paradigms. Figure \ref{forms} shows an example automaton 
for three forms. We refrain from describing a final automaton operation called Bounded
Local Optimization in \cite{walther:99} that was put to use here to
filter harmless spurious ambiguities from the original
version of fig.\/ \ref{forms}. The kind of ambiguity involved in our
Temiar grammar is one of alternative distribution of  technical symbols in strings of the same
segmental-content yield. Suffice to say that a simple parametrization
of Bounded Local Optimization, which could only look at \mbox{length-1}
transition paths emerging from any given state, was able to prune the
unwanted alternatives by considering technical transitions costlier in 
weight than segmental transitions.

\section{\label{conc}Conclusion}
The present paper has provided further support for \cite{walther:99}'s 
finite-state conception of One-Level Prosodic Morphology by formulating -- for the
first time -- a fully formalized and computational analysis of a complicated
piece of reduplicative morphology found in the Mon-Khmer language of
Temiar. Compared  to the initial proposal, all three core components
of enriched representations, namely technical transitions for
repeating or skipping segmental symbols and the ability to perform
infixation by using self loops, were again found necessary in the
course of this analysis. However, in Temiar the last 
enrichment -- $add\_self\_loops$ -- needed to be parametrized for a
prosodic condition to narrow down the insertion site to a unique
position per base. 

The prosodic condition of `stressed onset' proved \finallongpage
crucial to define that position, and accounted for the variation
between infixing aspectual morphology in longer bases and
descriptively prefixing morphology in monosyllabic ones. Temiar thus
underscores the utility of computing with real prosodic information in
finite-state morphology, a frequently missing desideratum according to \cite[p.170]{sproat:92}.
Also, the symmetry
of having both forward and backward-pointing technical transitions in
enriched automata representations was exploited in a  novel regular expression
operator called \code{seek(X)}, which encapsulated an interesting kind 
of ambiguous directional movement (or: movement underspecified for
direction)  towards a position satisfying property
\code{X}. This operator could rather directly be motivated from the data. In
particular, it facilitated an insightful account of the
base-length-dependent triggering of reduplication in the active
simulfactive aspect. 

Finally, in contrast to even the most recent analyses in
the theoretical linguistic literature, the full paradigm including the
causative forms was captured in this fairly complete analysis,
together with phonological modifications that sometimes occur between base and reduplicant, 
as exemplified by \l{yE}{\bf \l{m}}\l{yaa}{\bf \l{p}}. Apart from an
optional filtering step for some technical spurious ambiguities that
could make use of local optimization, neither global optimization nor
violable or soft constraints of the type argued for in Optimality
Theory \cite{prince.smolensky:93} were found necessary.

For future research, the empirical base of Temiar should be broadened
to include further reduplication patterns, in particular those found
in expressives. Also, the grammar should be amended to allow for words
containing geminates, which were initially excluded to simplify the
overall analysis at the cost of what is at best a peripheral aspect of it. Because the
finite-state constraints employed in this work are all surface-true,
the potential of machine-learning techniques to acquire them
automatically from surface-oriented corpora should be
explored. Finally, it would be very interesting to broaden to Temiar \finallongpage
the ongoing experiments with efficiency-oriented computational
variants of the One-Level Prosodic Morphology framework that were already alluded to in the
text. 

{\small


\begin{thebibliography}{}

\bibitem[\protect\citename{Antworth}1990]{antworth:90}
Evan Antworth.
\newblock 1990.
\newblock {\em PC-KIMMO: A Two-Level Processor for Morphological Analysis}.
\newblock SIL, Dallas.

\bibitem[\protect\citename{Beesley}1998]{beesley:98}
Kenneth~R. Beesley.
\newblock 1998.
\newblock Constraining separated morphotactic dependencies in finite-state
  grammars.
\newblock In {\em Proceedings of FSMNLP'98, Bilkent University, Turkey}, pages
  118--127.

\bibitem[\protect\citename{Benjamin}1976]{benjamin:76}
Geoffrey Benjamin.
\newblock 1976.
\newblock An outline of {T}emiar grammar.
\newblock In Philip Jenner, Lawrence Thompson, and Stanley Starosta, editors,
  {\em Austroastiatic studies}, volume~II, pages 129--187. University Press of
  Hawaii, Honululu.

\bibitem[\protect\citename{Bird and Ellison}1992]{bird.ellison:92}
Steven Bird and T.~Mark Ellison.
\newblock 1992.
\newblock {O}ne-{L}evel {P}honology: Autosegmental representations and rules as
  finite-state automata.
\newblock Technical report, Centre for Cognitive Science, University of
  Edinburgh.
\newblock EUCCS/RP-51.

\bibitem[\protect\citename{Bird and Ellison}1994]{bird.ellison:94}
Steven Bird and T.~Mark Ellison.
\newblock 1994.
\newblock {O}ne-{L}evel {P}honology.
\newblock {\em Computational Linguistics}, 20(1):55--90.

\bibitem[\protect\citename{Broselow and McCarthy}1983]{broselow.mccarthy:83}
Ellen Broselow and John McCarthy.
\newblock 1983.
\newblock A theory of infixing reduplication.
\newblock {\em The Linguistic Review}, 3:25--98.

\bibitem[\protect\citename{Gafos}1995]{gafos:95}
Adamantios Gafos.
\newblock 1995.
\newblock {O}n the {P}roper {C}haracterization of `{N}onconcatenative'
  {L}anguages.
\newblock Ms., Department of Cognitive Science, The Johns Hopkins University,
  Baltimore. (ROA-106 at http://ruccs.rutgers.edu/roa.html).

\bibitem[\protect\citename{Gafos}1996]{gafos:96}
Diamandis Gafos.
\newblock 1996.
\newblock {\em The articulatory basis of locality in phonology}.
\newblock {Ph.D.} thesis, The Johns Hopkins University, Baltimore, Md.
\newblock [Published by Garland:New York].

\bibitem[\protect\citename{Gafos}1998a]{gafos:98b}
Diamandis Gafos.
\newblock 1998a.
\newblock A-templatic reduplication.
\newblock {\em Linguistic Inquiry}, 29(3):515--527.

\bibitem[\protect\citename{Gafos}1998b]{gafos:98a}
Diamandis Gafos.
\newblock 1998b.
\newblock Eliminating long distance consonantal spreading.
\newblock {\em Natural Language and Linguistic Theory}, 16(2):223--278.

\bibitem[\protect\citename{Kaplan and Kay}1994]{kaplan.kay:94}
Ron Kaplan and Martin Kay.
\newblock 1994.
\newblock Regular models of phonological rule systems.
\newblock {\em Computational Linguistics}, 20(3):331--78.

\bibitem[\protect\citename{McCarthy}1982]{mccarthy:82a}
John McCarthy.
\newblock 1982.
\newblock Prosodic templates, morphemic templates, and morphemic tiers.
\newblock In Harry {van der Hulst} and Norval Smith, editors, {\em The
  structure of phonological representations, part I}, pages 191--224. Foris,
  Dordrecht.

\bibitem[\protect\citename{Prince and Smolensky}1993]{prince.smolensky:93}
Alan Prince and Paul Smolensky.
\newblock 1993.
\newblock Optimality theory. constraint interaction in generative grammar.
\newblock Technical Report RuCCS TR-2, Rutgers University Center for Cognitive
  Science.

\bibitem[\protect\citename{Shaw}1993]{shaw:93}
Patricia Shaw.
\newblock 1993.
\newblock The prosodic constituency of minor syllables.
\newblock In {\em Proceedings of the Eleventh West Coast Conference on Formal
  Linguistics}, pages 117--132, Stanford, CA. CSLI Publications.
\newblock [Distributed by Cambridge University Press].

\bibitem[\protect\citename{Sloan}1988]{sloan:88}
Kelly Sloan.
\newblock 1988.
\newblock Bare-consonant reduplication.
\newblock In {\em Proceedings of the Seventh West Coast Conference on Formal
  Linguistics}, pages 319--330, Stanford, CA. CSLI Publications.
\newblock [Distributed by Cambridge University Press].

\bibitem[\protect\citename{Sproat}1992]{sproat:92}
Richard Sproat.
\newblock 1992.
\newblock {\em Morphology and Computation}.
\newblock MIT Press, Cambridge, Mass.

\bibitem[\protect\citename{van Noord}1997]{vannoord:97}
Gertjan van Noord.
\newblock 1997.
\newblock {FSA} {U}tilities: A toolbox to manipulate finite-state automata.
\newblock In Darrell Raymond, Derrick Wood, and Sheng Yu, editors, {\em
  Automata Implementation}, volume 1260 of {\em Lecture Notes in Computer
  Science}, pages 87--108. Springer Verlag.
\newblock (Software under {\footnotesize\tt
  http://grid.let.rug.nl/$\sim$vannoord/Fsa/}).

\bibitem[\protect\citename{Walther}1999a]{walther:99a}
Markus Walther.
\newblock 1999a.
\newblock {\em Deklarative prosodische Morphologie: constraint-basierte
  {A}nalysen und {C}omputermodelle zum {F}innischen und {T}igrinya}.
\newblock Niemeyer, T{\"u}bingen.

\bibitem[\protect\citename{Walther}1999b]{walther:99}
Markus Walther.
\newblock 1999b.
\newblock One-{L}evel {P}rosodic {M}orphology.
\newblock Marburger Arbeiten zur Linguistik~1, University of Marburg.
\newblock 64 pp. \\ (http://xxx.lanl.gov/abs/cs.CL/9911011).

\bibitem[\protect\citename{Walther}2000]{walther:00}
Markus Walther.
\newblock 2000.
\newblock Finite-state {R}eduplication in {O}ne-{L}evel {P}rosodic
  {M}orphology.
\newblock In {\em Proceedings of NAACL-2000}, pages 296--302, Seattle/WA. North
  American Association for Computational Linguistics, Morgan Kaufman.
\newblock (http://xxx.lanl.gov/abs/cs.CL/0005025).

\end{thebibliography}
\end{document}